\documentclass[preprint,12pts,doublespacing]{elsarticle}

\usepackage{amssymb}
\usepackage{graphicx}
\usepackage{subfigure}
\usepackage{amsmath}
\usepackage{mathrsfs} 
\usepackage{xcolor}
\usepackage{booktabs}
\usepackage{multirow}

\usepackage{pdflscape}
 \usepackage{comment}
 \usepackage{color}
 \usepackage{url}

\newcommand{\toadd}[1]{{#1}}
\newcommand{\revisiontwo}[1]{{{#1}}}

\journal{XXXXX}

\begin{document}

\begin{frontmatter}

\title{\toadd{Do Deep Neural Networks Contribute to Multivariate Time Series Anomaly Detection?}}

\author[inst1,inst2]{Julien Audibert\corref{cor1}}
\ead{julien.audibert@eurecom.fr}

\affiliation[inst1]{organization={Orange},
            city={Sophia Antipolis},
            country={France}}

\author[inst2]{Pietro Michiardi}
\ead{pietro.michiardi@eurecom.fr}
\author[inst3]{Fr\'{e}d\'{e}ric Guyard}
\ead{frederic.guyard@orange.com}
\author[inst1]{S\'{e}bastien Marti}
\ead{sebastien.marti@orange.com}
\author[inst2]{Maria A. Zuluaga\corref{cor1}}
\ead{zuluaga@eurecom.fr}

\affiliation[inst2]{organization={EURECOM},
            city={Sophia Antipolis},
            country={France}}

\affiliation[inst3]{organization={Orange Labs},
            city={Sophia Antipolis},
            country={France}}
\cortext[cor1]{Corresponding author}

\begin{abstract}
Anomaly detection in time series is a complex task that has been widely studied. In recent years, the ability of unsupervised anomaly detection algorithms has received much attention. This trend has led researchers to compare only learning-based methods in their articles, abandoning some more conventional approaches. As a result, the community in this field has been encouraged to propose increasingly complex learning-based models mainly based on deep neural networks. To our knowledge, there are no comparative studies between conventional, machine learning-based and, deep neural network methods for the detection of anomalies in multivariate time series.  In this work, we study the anomaly detection performance of \revisiontwo{sixteen} conventional, machine learning-based and, deep neural network approaches on five real-world open datasets. By analyzing and comparing the performance of each of the \revisiontwo{sixteen} methods, we show that no family of methods outperforms the others. 
Therefore, we encourage the community to reincorporate the three categories of methods in the anomaly detection in multivariate time series benchmarks.  
\end{abstract}


\end{frontmatter}

\section{Introduction}
 A multivariate time series is a set of measurements correlated with each other over time, which models and represents the behavior of a multivariate process in time. Multivariate time series are used in a large number of fields such as industrial control systems \cite{USAD}, finance \cite{finance}, and healthcare \cite{health}. 
 Detecting unexpected behavior or patterns that do not conform to the expected behavior of previously seen data is an active research discipline called anomaly detection in multivariate time series \cite{chandola2009,domingues2018}. Anomalies may indicate a significant problem in several applications. For example, an anomaly in industrial control systems may indicate a malfunction, financial anomalies may be the result of a fraud, or they may indicate diseases in healthcare. Being a critical task, there is a wide range of methods that have been developed to address it~\cite{domingues2018}.

Over the last decade, there has been an increased enthusiasm around deep neural networks (DNNs)~\cite{DNN} thanks to a reported number of successes in multiple tasks \cite{finance, health, Zong2018}, and their demonstrated ability to infer high-order correlations in complex data with potentially large volume and dimensionality. Multivariate time series anomaly detection has been no exception to the trend, leading to an explosion of DNN-based methods suggesting methodological advances and improved performance, as presented in~\cite{audibertfigshare}. DNN-based methods aim to learn deep latent representations of the multivariate time series to infer a model of variability, that is then used for anomaly grading in unseen data. The rationale behind the increased use of DNN architectures lies in the need of learning potentially complex data patterns underlying the temporal evolution of multivariate data.

Under the above-mentioned argument and motivated by the good performance of DNNs in other fields, researchers have moved away from comparisons with more traditional methods, i.e. machine learning and conventional/statistical methods (e.g. \cite{USAD,Su2019,Zong2018}). % \cite{hodge2004} 
This trend has encouraged the community to develop even more complex models to improve the performance of DNN-based methods, without any theoretical or empirical evidence that these are superior to the more established body of methods in the literature.

DNN-based models are complex to train, involving the estimation of a large amount of parameters and requiring large training sample sizes and computational resources. Moreover, their complexity continues to grow as larger models continue to be developed. Instead, conventional models are simpler, lighter, easier to interpret, and often better adapted to the constraints of real-world applications. It is therefore crucial to determine if the complexity brought in by DNN-based methods is a necessary price to pay for a gain in performance or if the progress reported in recent years is illusory \cite{eamonn} and the use of conventional methods should be preferred. The lack of a general comparison covering all families of methods does not allow to answer this question and hinders the translation and use of DNN-based methods in real-world applications.
 
In this paper, we aim to close this gap by establishing a thorough comparison between conventional methods, machine learning-based and more recent DNN-based approaches. Our work is motivated by different recent works, which have reported on the limitations and drawbacks of DNN-based methods in different application fields~\cite{antun2020,heaven2019,jiao2020,makridakis2018_2}. While some of this works have focused on pointing out to the weaknesses of DNN-based methods~\cite{antun2020,heaven2019}, other works have been able to demonstrate the superiority of more conventional approaches~\cite{jiao2020,makridakis2018_2}. 

The main contributions of this paper are the following:  
\begin{itemize}

    \item We describe \revisiontwo{sixteen} of the most commonly used methods for anomaly detection in multivariate time series grouped into three categories: Conventional, machine learning-based and DNN-based. 
    \item We study and analyze the performance of these techniques over five open real-world data sets. 
    \item Finally, we discuss the need for DNN-based approaches and the importance of conventional methods in future benchmarks for multivariate time series anomaly detection.   
\end{itemize}

The rest of this document is organized as follows. Section~\ref{sec:related} briefly reviews other works comparing modern DNN-based methods to previous non-DNN-based works. Section \ref{sec:def} formalizes the problem of anomaly detection in multivariate time series. Section \ref{sec:methods} presents the methods compared in this study.   Sections \ref{sec:experiments} describe the experiments and analyze the performance on the data sets.

\section{Related Works}~\label{sec:related}
Different studies have raised the question about the real gain of DNN-based methods in several application fields. 

A first study by Jiao \textit{et al.}~\cite{jiao2020},  showed how conventional linear regression methods outperform DNN-based techniques in two showcased optical imaging problems, i.e. an optical cryptosystem attack and  blind reconstruction in single pixel imaging. Autun \textit{et al.}~\cite{antun2020} proposed a stability test to demonstrate how DNN-based methods for image reconstruction are very sensitive to tiny perturbations in the input images during training, which leads to unstable results. Furthermore, Heaven~\cite{heaven2019} showed small changes in a DNN's input, usually imperceptible to humans, can destabilize the best neural networks, thus pointing to the lack of robustness of DNN-based methods and their dependence on large amounts of data. Most recently, in the context of medical image segmentation, Fu \textit{et al.}~\cite{fu2021} showed that simpler DNN configurations have better generalization properties than complex state-of-the-art DNN models, thus challenging the current trend towards continuously increasing the model complexity.

In the specific context of time series analysis, Garg\textit{et al.}~\cite{garg2021} raise the fact that there is no benchmark in the literature between multivariate methods for anomaly detection in time series. \toadd{They propose} a benchmark \toadd{of 15 methods out of which 12 are DNNs methods.} 

Finally, the results of the M3 challenge on time series forecasting~\cite{makridakis2018_2} showed that the accuracy of machine learning and DNN models, in general, was lower than that one of conventional approaches, while their computational requirements were considerably greater than those of conventional statistical methods. Similarly, one of the main outcomes of the follow-up M4 competition~\cite{makridakis2020} was that none of the pure ML methods participating was able to outperform the combination of DNN models and statistical (i.e. conventional) methods. For instance, only one DNN approach was more accurate than a naïve random walk model that assumed future values will be the same as those of the last known observation~\cite{makridakis2018_2}.

\section{Definitions}~\label{sec:def}
This section is concerned with providing a clear definition of multivariate time series (Section \ref{sec:mts}), of the types of anomalies that can be found in a multivariate time series (Section \ref{sec:Types}), and of formalizing the problem of anomaly detection in multivariate time series (Section \ref{sec:Formulation})

\subsection{Multivariate Time Series}~\label{sec:mts}
A multivariate time series is a sequence of data points $\mathcal{T}=\{\mathbf{x}_1, \ldots ,\mathbf{x}_T \}$, $\mathbf{x}_t\in \mathbb{R}^m \forall \,\, t$, each being an observation measured at a specific time $t$ in a process, with $m$ the number of time-series. A  univariate  time series is a special case of multivariate time series with $m=1$. In this paper we focus on the general context, multivariate time series.

\subsection{Types of Anomalies}~\label{sec:Types}
Anomalies in time series, also called outliers, are points or sequences of points that do not correspond to normal behavior~\cite{chandola2009}. The concept of normal behaviour is difficult to formalize. Therefore, another possible definition for anomalies could be a pattern in data that is not expected in comparison to what has been seen before \cite{chandola2009}. In fact, an implicit assumption is that anomalies are rare events. 
%Anomalies should not be confused with the noise present in the time series. Noise is a phenomenon which, unlike anomalies, has less interest in being analyzed.  

Chandola et al.~\cite{chandola2009} propose to classify time series anomalies into three types, point, contextual and collective anomalies. 

\begin{figure*}
\centering
\includegraphics[scale=0.46]{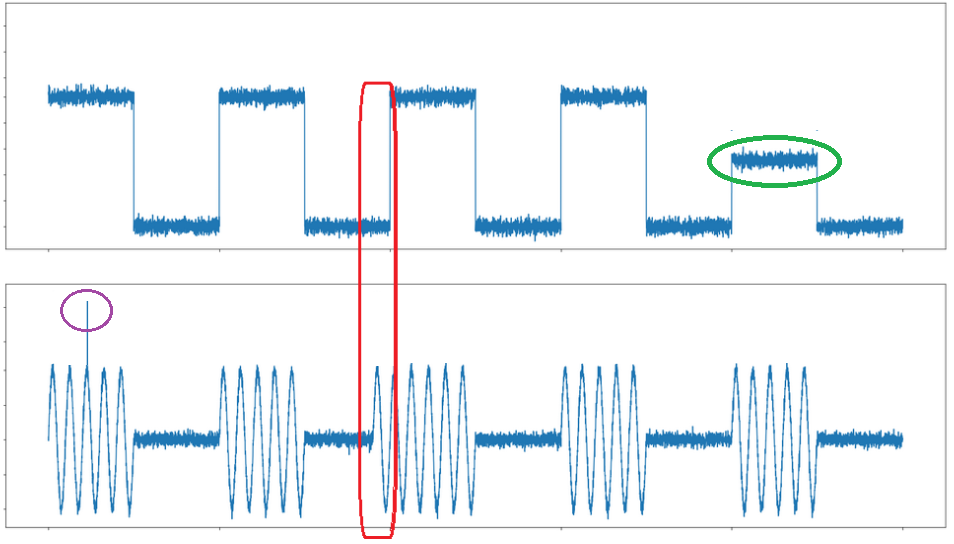}
\vspace{-1.5em}
\caption{An example of a point anomaly (in purple), a contextual anomaly (in red) and a collective anomaly (in green)}
\vspace{-0.5em}
\label{fig:anomalies}
\end{figure*}

\begin{itemize}
    \item \textbf{Point anomalies.} This is the simplest type of anomaly. It corresponds to a point that differs from the rest of the data (Figure \ref{fig:anomalies}).
    
    \item \textbf{Contextual anomalies.} A contextual anomaly can be defined as follows : A data point is anomalous in a specific context, but is considered normal in another context, meaning that observing the same point through different contexts does not always give a sign of abnormal behavior. For example, a temperature of 30°C during summer is normal, while the same temperature in winter is abnormal. Figure \ref{fig:anomalies} illustrates a contextual anomaly, where the values of the two time-series are not abnormal taken individually, but, seen in context, the values of the bottom time-series should be close to 0.
    
    \item \textbf{Collective anomalies} A collective anomaly corresponds to a group of anomaly points. Each individual point might not be an anomaly, but their appearance together is an anomaly (Figure \ref{fig:anomalies}).
    
\end{itemize}

\subsection{Problem Formulation}~\label{sec:Formulation}
Let us now consider a problem where $\mathcal{T}$ is given, and let us introduce a binary variable $y\in\{0,1\}$. The goal of anomaly detection methods is to learn a model that assigns to an observation $\mathbf{\hat{x}}_t, \,t>T$ , a label $y_t$ to indicate a detected anomaly at time $t$, \textit{i.e.} $y_t=1$, or not ($y_t=0$). 

Supervised anomaly detection assumes that, jointly with $\mathcal{T}$, there are labels that assign to every $\mathbf{x}_t \in \mathcal{T}$ one of two classes, normal or abnormal. During training, the model is thus expected to learn how to classify anomalies by using both $\mathcal{T}$ and the associated labels. 

Unsupervised anomaly detection assumes $\mathcal{T}$ contains only normal points. Therefore, the model is trained to learn the distribution of normal data and an anomalous point is expected to be one that differs significantly from $\mathcal{T}$. 
The difference between the sample $\mathbf{\hat{x}}_t$ and the normal set $\mathcal{T}$ is measured by an anomaly score, which is then compared to a threshold. If the score is above the threshold the point is considered as an anomaly.  

In practice, it is difficult to obtain sets $\mathcal{T}$ with a sufficient number of samples from both classes, a requirement of supervised techniques. As a result, unsupervised methods are the most common approach found in the literature~\cite{chandola2009}. Therefore, this study focuses on unsupervised techniques. 

\section{Anomaly detection methods}~\label{sec:methods}
In this section, we first present the different categories to classify different anomaly detection methods (Section~\ref{sec:categories}). Next, we describe the \revisiontwo{sixteen} multivariate time series anomaly detection methods used in our experiments. Each method is presented within one of three categories that we propose: conventional (Section \ref{sec:conventional}), machine learning-based (Section \ref{sec:ml}) and DNN-based  methods (Section \ref{sec:dnn}).

\subsection{Taxonomy}~\label{sec:categories}
We define a taxonomy consisting of three classes of anomaly detection methods for multivariate time series. These are: conventional approaches, machine learning-based and DNN-based methods.

\textbf{Conventional approaches}, rely on the assumption that the observed data is generated by a stochastic model and their aim is to estimate a model's parameters from the data and then use the model for prediction~\cite{breiman}. It is often the case that the model hypothesis is considered linear.

The boundary between conventional and machine learning-based approa\-ches is not fully clear. \textbf{Machine learning-based} models produce predictions about the results of complex mechanisms by mining databases of inputs for a given problem, without necessarily having an explicit assumption about a model's hypothesis. In this setup, a method aims to learn a function that operates an input data to predict output responses~\cite{breiman}. 

Finally, \textbf{DNN-based methods} are a subclass of non-linear machine learning-based methods which use neural networks with multiple layers \cite{DNN}. 

\subsection{Conventional methods}~\label{sec:conventional}
The methods presented in this section offer many different approaches. We have classified them into five categories. Control chart, where the objective is to monitor variations in the statistical characteristics of a process. Prediction methods, where the objective is to predict the next point. Decomposition techniques, based on the search for unusual patterns in time series using decomposition and finally similarity-search model based on the search for similar sub-sequences in the data.

\subsubsection{Control Charts methods} ~\label{sec:ccmethods}
Control charts are a statistical process control tool. Their goal is to track the mean and variance of a time series over time and to detect abrupt changes in its statistical behaviour.
We consider two methods: Multivariate CUmulated SUMs control chart (MCUSUM) and Multivariate Exponential Weighted Moving Average (MEWMA).

In \cite{MCUSUM}, Woodall and Ncube proposed to monitor the performance of multiple variables in a manufacturing system using a control chart called \textbf{Multivariate CUmulated SUMs control chart} \textbf{(MCUSUM)}. The MCUSUM uses the fact that the cumulative sum $St$ of the recursive residuals of a statistic $s$ compared to a normal value (like the mean) : $St = s(\mathbf{x}_{t-k+1},\ldots, \mathbf{x}_t)$ is stable for normal data and increasing after the change. Using this property, the MCUSUM anomaly score $g_t$ is based on the comparison of the increase of $S_t$ with a threshold
$h$: $g_t = max (0, St - \mu  + g_{t-1})$ with $g_0 = 0$ and $\mu$ the value of $St$ for normal data. MCUSUM iterates over $g_t$ as long as $g_t < h$. If $g_t \geq h$, an alarm is raised and the iteration over $g_t$ is restarted with $g_0 = 0$ in order to detect another change point. Due to this sequential restarting of the test, MCUSUM detects small but persistent structural changes.

\textbf{Multivariate Exponential Weighted Moving Average }\textbf{(MEWMA)} \cite{MEWMA} is based, as its name indicates, on the exponential smoothing of data. For a given constant $\lambda \in [0,1]$, the successive data are smoothed using: $g_t = \lambda \mathbf{x}_t+(1-\lambda)g_{t-1}$. Unlike MCUSUM, MEWMA gives more importance to recent history values.

\subsubsection{Forecast methods} 
Control Charts methods are usually based on the hypothesis that the individual data in the time-series are independent and identically distributed. This assumption is rarely satisfied in practice. A generic way to handle this issue is to built a mathematical model, incorporating the known or assumed sequential correlations of the time-series. Using this model, the value $\mathbf{x}_t$ of the data indicator at $t$ is expressed as $\mathbf{x}_t = z_t + e_t$, where $z_t$ is accounting for normal sequential correlations within the data and $e_t$ (the residual) is the noise.
Once a mathematical model representing time-series evolution is chosen, a usual approach is to predict at time $t$ the expected value $\mathbf{\hat{x}}_{t + 1}$. The anomaly score can then be expressed as the difference between $\mathbf{\hat{x}}_{t+1}$ and the actual value $\mathbf{x}_{t + 1}$. 

The most commonly used forecast method in multivariate time series is the \textbf{Vector Autoregressive (VAR)} method. VAR is a statistical model that captures the inter-dependencies between several time series.
In a VAR model, variables are treated symmetrically so that each variable is explained by its own past values and by the past values of the other variables. For example, in \cite{VAR2}, VAR is used to monitor multivariate processes.

\subsubsection{Decomposition methods} 
Methods in this category use basis functions for decomposing the time series. Given $\mathbf{x}_t$, a point of a multivariate time series, it can be expanded in terms of the eigenfunctions $\phi_{j}$ as: $
\mathbf{x}_t = \sum_{j=1}^{\infty}\alpha_{j} \phi_{jt}$, where the coefficients  $\alpha_{j}$ are given by the projection of $\mathbf{x}_t$ on the respective eigenfunctions.

Anomaly detection is then performed by computing the projection of each new point onto the eigenvectors, and a normalized reconstruction error. The normalized error is used as the anomaly score.
We discuss three methods under this sub-category: Principal Component Analysis (PCA), Singular Spectrum Analysis (SSA) and Independent Component Analysis (ICA).

 \textbf{Principal Component Analysis (PCA).}  
The idea of PCA\cite{PCA} is to group data points into clusters defined as an ellipsoid in $\mathbb{R}^m$ containing the normal data. The idea is that the length of the principal axes of the ellipsoid represents the direction of the variability of the data instances.

The (always positive) eigenvalues $\lambda_i$ with $i = 1,\ldots, k$ corresponding to the principal component $v_i$ characterize the variability in the dataset captured by the eigenvector $v_i$. A subspace $S \subset \mathbb{R}^m$ is selected using the $r$ eigenvectors $v_i$ corresponding the $r$ largest eigenvalues.

In \cite{PCA_anomaly}, the Q-statistic that characterizes the PCA projection residual statistics is used to define the threshold $h_{\alpha}$, such that the anomaly score is: $
   \left | \widetilde{\mathbf{x}_{t}} \right | > h_{\alpha} $ then $\mathbf{x}_{t}$ is an anomaly with $ 1-\alpha $ confidence where $\alpha \in [0,1]$ is a manually defined parameter.

\textbf{Singular Spectrum Analysis (SSA)} considers a time series as a projection of the trajectory of a dynamical system in a vector space $V = \mathbb{R}^m$ where $\mathbf{x}_{t}$ is the position at time $t$ of a state of the system. From dynamical systems theory, it is known that physically observable states of a dynamical system in $\mathbb{R}^m$ are lying on attractors of the dynamics (i.e. subsets of the $\mathbb{R}^m$ capturing all the long term evolution of the system). Future observable states should normally continue to be located on these attractors. As such, $\mathbf{\hat{x}}_{t}$ are assumed to be located on (or close to) an attractor of the dynamics.
So, the distance between $\mathbf{\hat{x}}_{t}$ and the attractor computed at time $t$ is evaluated and used as anomaly score \cite{SSA_TS}.

\textbf{Independent Component Analysis (ICA)} \cite{ICA} assumes that the different physical processes that generate multivariate time series are statistically independent of each other. % 

ICA decomposes a multivariate time series into ``independent" components by orthogonal rotation and by maximizing the statistical independence between the components assuming their non-Gaussianity. In \cite{ICA_anomaly}, the authors use Kurtosis, a classical measure of non-Gaussianity, as an anomaly score. A negative Kurtosis indicates a uniformly distributed factor, indicating a clustered structure and a high positive Kurtosis identifies a multivariate anomaly.

\subsubsection{Similarity-search approach}
Similarity search methods \cite{Xia1997} are designed to identify patterns in a multivariate time series. These methods compute the distance between all the sub-sequences of the time series. The minimum distance between a sub-sequence and all others is used as an anomaly score. 

The \textbf{Matrix-Profile (MP)} \cite{MP} is a data structure for time series analysis. 
It consists of three elements. The first element is the distance profile, which is a vector of Euclidean distances between a given sub-sequence $W$ and each sub-sequence in the set $A$ of all sub-sequences of the multivariate time series. The distance is measured using the z-normalized Euclidean distance between sub-sequences. 
The distance profiles are arranged into a data structure denoted the distance matrix, which corresponds to the second element in MP.
\toadd{The distance matrix corresponds to all the distance profiles that have been computed for each reference sub-sequence. 
At the same time, the matrix profile is the simplification of the distance matrix by considering only the nearest neighbor for each sub-sequence allowing a memory complexity of $O(|A|)$.} The vector obtained corresponds to the smallest values of each row in the matrix. It is defined as : $MP(i) = min(d(W^i,W^j))$ with $W^j \in  A \setminus{\{W^i\}}$.

A low value in the matrix profile indicates that the sub-sequence has at least one relatively similar sub-sequence located somewhere in the original series. In \cite{MP-MAD}, it is shown that a high value indicates that the original series must have an abnormal sub-sequence. Therefore the matrix profile can be used as an anomaly score, with a high value indicating an anomaly.

\subsection{Machine learning-based methods}~\label{sec:ml}
The methods presented in this section fall into three categories :  Isolation, Neighbourhood-based and Domain-based methods, which have been proposed in the survey by Domingues \textit{et al.}~\cite{domingues2018}.  
Isolation algorithms consider a point as an anomaly when it is easy to isolate from others. Neighbourhood-based models look at the neighbourhood of each data point to identify outliers. Domain-based methods rely on the construction of a boundary separating the nominal data from the rest of the input space. 

A common characteristic of machine learning-based techniques is that they typically model the dependency between a current time point and previous ones by transforming the multivariate time series $\mathcal{T}$ into a sequence of windows $\mathcal{W}=\{W_1, \ldots, W_T\}$, where $W_t$ is a time window of length $K$ at a given time $t$ : $
    W_t=\{\mathbf{x}_{t-K+1}, \ldots,\mathbf{x}_{t-1},  \mathbf{x}_{t}\}$.  
In this setup, learning-based anomaly detection methods assign to a window $\widehat{W}_t, \, t> T$, a label $y_t$ to indicate a detected anomaly at time $t$, \textit{i.e.} $y_t=1$, or not ($y_t=0$) based on the window's anomaly score. We will use this notation in the following.   

\subsubsection{Isolation methods}
Isolation methods focus on separating outliers from the rest of the data points. These methods attempt to isolate the anomalies rather than mapping the normal points.

The \textbf{Isolation Forest (IF) algorithm} \cite{IF} is based on decision trees. IF calculates, for each time window an anomaly score. To calculate this score, the algorithm isolates the sample recursively: it chooses a feature and a ``cut-off point" at random, then evaluates whether this isolates the sample; if so, the algorithm stops, otherwise, it chooses another feature and another cut-off point at random, and so on until the data is isolated from the rest. %\cite{IF,IF_MTS}
The number of  cut-off point defines the anomaly score: a sample with a very short path, i.e. sample that is easy to isolate, is also likely to be an anomaly since it is very far from the other samples in the dataset.

\subsubsection{Neighbourhood-based methods}
Among neighborhood-based methods, which study the neighborhoods of every point to identify anomalies, the \textbf{Local Outlier Factor (LOF)} \cite{LOF} measures the local deviation of a given data point with respect to its neighbours. Based on the  K-nearest neighbors, the local density of an observation is evaluated by considering the closest K observations in its neighbourhood. The anomaly score is then calculated by contrasting its local density with those of its k-nearest neighbors. A high score indicates a lower density than its neighbors and therefore potentially an anomaly. It has been applied to multivariate time series \cite{LOF_MTS}, demonstrating its ability to detect anomalies in long-term data.

\textbf{Density-based spatial clustering of applications with noise (DBSCAN)} \cite{DBSCAN} is a clustering method that 
groups data points in high density areas (many nearby neighbors) and marks points in low-density regions (few neighbors) as anomalous. To handle multivariate time series, DBSCAN considers each time window as a point with the anomaly score being the distance from the point to the nearest cluster.%\cite{DBSCAN_MTS}

\subsubsection{Domain-based methods}
Domain-based methods aim to construct a boundary between normal samples and the rest of the input space. The distance of a points to this boundary is used as the anomaly score. Among these, the \textbf{One-Class Support Vector Machine (OC-SVM)} \cite{OC-SVM} method learns  the smallest hypersphere  containing of all the training data points. The learned model classifies points inside the hypersphere as normal and labels those in the rest of the space as anomalous. An anomaly score can also be obtained by taking the signed distance from the hyper-sphere. The signed distance is positive for a normal value and negative for an abnormal.
As other machine learning methods One-class SVM has been used for time series anomaly detection by using time windows rather than the raw series. J.Ma and S.Perkins \cite{OC-CSM_MTS} use this approach, while proposing to combine the one-class SVM output for different time windows to produce more robust detection results.

\subsection{DNN-based methods}~\label{sec:dnn}
DNN-based methods are a are sub-category of machine learning-based approaches, which rely on deep neural networks. Given the explosion of DNN-based methods over the last years, we present them as a separate category. We selected the \revisiontwo{five} methods that report a good performance in the literature on the five datasets used in this paper.
These are: Auto-Encoder (AE), %\revisiontwo{Univariate Fully-Connected Auto-Encoder with Dynamic Gaussian scoring (UAE with Gauss-D)}, 
UnSupervised Anomaly Detection (USAD), Long Short-Term Memory Variational Auto-Encoders (LSTM-VAE), Deep Autoencoding Gaussian Mixture Model (DAGMM) and OmniAnomaly (OA).

An \textbf{Auto-Encoder (AE)} ~\cite{Autoencoder} is a neural network architecture consisting of a combined encoder and decoder. The encoder maps the input windows into a set of latent variables, while the decoder maps the latent variables back into the input space as a reconstruction. The difference between the input window and its reconstruction is the reconstruction error. The AE learns to minimize this error. The anomaly score of a window is the corresponding reconstruction error. A window with a high score is considered abnormal. 

% \revisiontwo{The \textbf{Univariate Fully-Connected Auto-Encoder with Dynamic Gaussian scoring} (UAE with Gauss-D)~\cite{garg2021} trains a separate Auto-Encoder for each dimension. Then it fits a  Gaussian distribution, with dynamic mean and variance, to the training errors and design a score based on the fitted distribution. }

\textbf{UnSupervised Anomaly Detection (USAD)}~\cite{USAD} extends the AE concept and constructs two AEs sharing the same encoder. The architecture is driven in two phases. In the first phase, the two AEs learn to reproduce the normal windows. TIn the second phase, an adversarial training teaches the first AE to fool the second one, while the second one learns to recognize the data coming from the input or the reconstructed by the first AE. The anomaly score is the difference between the input data and the data reconstructed by the concatenated AEs.

\textbf{Long Short-Term Memory Variational Auto-Encoders} \textbf{(LSTM-VAE)} ~\cite{lstm-vae} uses an LSTM to model temporal dependency, whereas the VAE projects the input data and its temporal dependencies into a latent space. During decoding, the latent space representation allows to estimate the output distribution. An anomaly is detected when the log-likelihood of the current data is below a threshold. S.Lin et al.~\cite{LSTM-MTS} have shown LSTM-VAE's capacity to identify anomalies that span over multiple time scales. % \cite{Lstm}

The \textbf{Deep Autoencoding Gaussian Mixture Model (DAGMM)} \cite{Zong2018} considers a Gaussian Mixture Model (GMM) and a Deep Auto-encoder together to model the distribution of multidimensional data. The purpose of the Deep Auto-encoder is to generate a low-dimensional representation and a reconstruction error for each input data time window. This representation is used as input of a GMM. The parameters of the Deep Auto-encoder and the mixture model are optimized simultaneously from end to end, taking advantage of a separate estimation network to facilitate the learning-based of the parameters of the mixture model.
The DAGMM then uses the likelihood to observe the input samples as an anomaly score. 

Finally, \textbf{OmniAnomaly (OA)} \cite{Su2019} is a recurrent neural network that learns robust representations of time series  with a planar normalizing flow and a stochastic variable connection. Then the reconstruction probabilities are used to determine anomalies. OmniAnomaly uses the Gated Recurrent Unit (GRU) to model the time dependencies of multivariate time series. The method also uses a VAE to learn a mapping of the input data $W$ to a representation in a latent space. To model time dependencies in the latent space, OmniAnomaly uses a stochastic variable connection technique. As suggested by \cite{Donut}, the reconstruction can be evaluated by a conditional probability. The anomaly score used is then the probability of reconstruction. A high score means that the input can be well reconstructed. If an observation follows normal time series patterns, it can be reconstructed with high confidence. On the other hand, the lower the score, the less well the observation can be reconstructed and the more likely it is to be anomalous.

\section{Experiments and results}~\label{sec:experiments}
We present the experimental setup (Sec.~\ref{sec:setup}). Then, we study the performance of the  \revisiontwo{sixteen} methods from the different categories (Sec.~\ref{sec:methods}) on the five benchmark datasets. We analyze these results by comparing the anomalies detected by the conventional methods, \toadd{the machine learning and the DNN-based approaches} (Sec. \ref{sec:analyze}). We investigate the impact of the size of the training set (Sec.~\ref{sec:size_train}). Finally, we summarize the results and discuss them (Sec.~\ref{sec:results}). 
\subsection{Experimental setup}~\label{sec:setup}
We first describe the benchmark datasets and the performance metrics used in our study.

\paragraph{Benchmark Datasets}~\label{sec:datasets}
Table~\ref{table:dataset} \toadd{summarizes the five datasets used in this study and reports the train and test splits used during the experiments}. The \textbf{SWaT} dataset\footnote{{https://itrust.sutd.edu.sg/itrust-labs\_datasets/dataset\_info/\#swat}} is a scaled down version of a real-world industrial water treatment plant. The \textbf{WADI} dataset\footnote{{https://itrust.sutd.edu.sg/itrust-labs\_datasets/dataset\_info/\#wadi}} is an extension of the SWaT testbed. \textbf{SMD} is a 5-week-long dataset from an Internet company made publicly available\footnote{{https://github.com/smallcowbaby/OmniAnomaly}}. \textbf{SMAP} and \textbf{MSL} are real-world public datasets from NASA\footnote{{https://s3-us-west-2.amazonaws.com/telemanom/data.zip}}. \toadd{ We highlight the fact that, while these datasets are among the most common benchmarks for multi-variate time-series anomaly detection validation, \textbf{ SMD}, \textbf{SMAP} and \textbf{MSL} datasets may contain flaws, as pointed out in~\cite{eamonn}.} 

\begin{table}[t] %[H]
 \caption{Benchmarked Datasets. (\%) is the percentage of anomalous data points in the dataset.}
  \label{table:dataset}
  \centering
  \begin{tabular}{lrrrcr}
    \toprule
    Dataset&Train&Test&Dimensions&Anomalies (\%)&Anomaly\\
    \midrule
    SWaT & 496800& 449919 & 51 & 11.98 & P, C, Col\\
    WADI & 1209601 & 172801 & 123  & 5.99 & P, C, Col\\
    SMD & 708405 & 708420 & 28*38 & 4.16 & P, Col\\
    SMAP & 135183 & 427617 & 55*25 & 13.13 & P, Col\\
    MSL &  58317 & 73729& 27*55 & 10.72 & P, Col\\
  \bottomrule
  \multicolumn{6}{l}{\small P: Point Anomaly, C: Contextual anomaly, Col: Collective anomaly}\\
\end{tabular}
\end{table}

\paragraph{Evaluation Metrics}
\toadd{
We use the F1 score (F1) and the average precision (AP) from anomaly scores to evaluate anomaly detection performance. The F1 is estimated as $F1 = 2 \cdot  P \cdot R/(P + R)$, with $P = (TP)/(TP+FP)$, the precision, $R = TP/(TP+FN)$, the recall, TP the True Positives window, FP the False Positives, and FN the False negatives.} In our experiments, we consider a window is labeled as an anomaly as soon as one of the points it contains is detected as anomalous. For methods that do not use a time window, an anomaly is considered detected if the detection occurs \toadd{within the length of sliding window time points on either side of the anomaly.}

\toadd{The AP summarizes the P-R curve as the weighted mean of precision achieved at every given anomaly score threshold, with the increase in recall from the previous threshold used as the weight. As such, it allows to obtain a score that takes into account the ability of the methods to rank the anomalies without having to define a threshold, thus providing a complementary information w.r.t. the F1 score. It is estimated as}: $   AP = \sum_n (R_n - R_{n-1}) P_n$ where $P_n$ and $R_n$ are the precision and recall at the n-th threshold. \toadd{ The AP is particularly sensitive to the positive class, i.e. the anomalies, so it is well-suited to highly unbalanced data, as it is the case in anomaly detection. }

\subsection{Benchmark Performance}\label{sec:performance}
\begin{figure*}[t]
\begin{tabular}{cc}
   \includegraphics[width=0.465\linewidth]{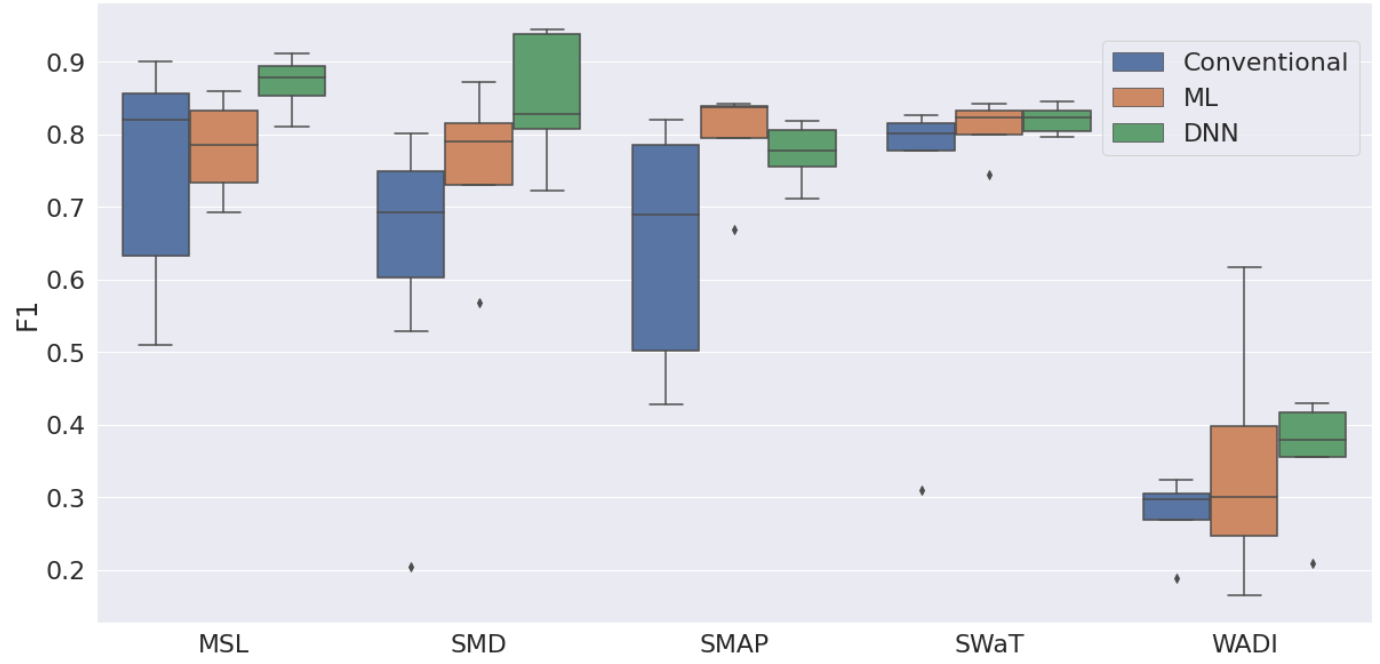}  &  \includegraphics[width=0.465\linewidth]{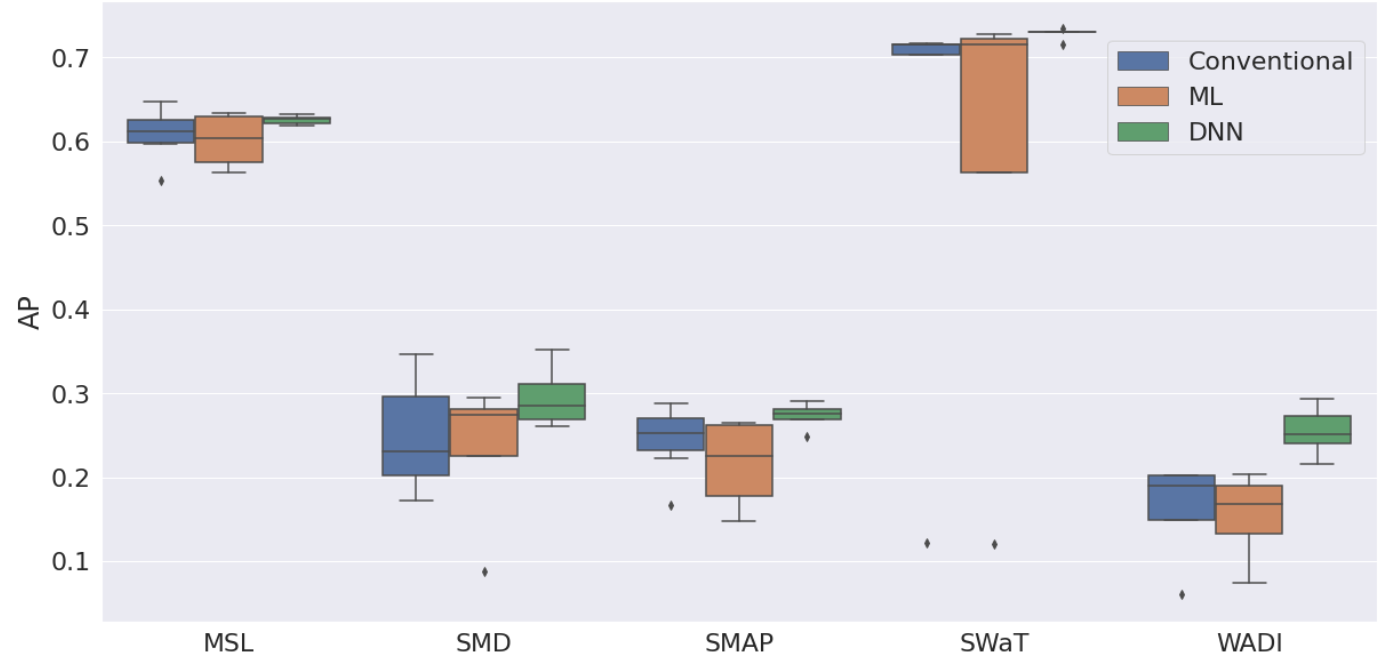}\\
\end{tabular}
\caption{\toadd{Boxplots of F1 (left) and AP (right) for each category of methods on the five benchmark dataset.}}
\label{fig:performance}
\end{figure*}

We assessed the performance of the \revisiontwo{sixteen} anomaly detection methods on the five benchmark datasets in terms of F1 score and AP. 
As not all of the anomaly detection approaches provide a mechanism for selecting anomaly thresholds, which are required for F1 estimation, we tested a thousand possible anomaly thresholds for each model. In particular, we normalized the anomaly score between 0 and 1 and then tested one thousand thresholds in steps of 0.001. We report the results associated to the highest achieved F1 score \toadd{among the thresholds. Figure \ref{fig:performance} presents boxplots summarizing the performance of each family of methods. Table~\ref{table:Summary} lists the best and worst performing method for each family of methods.}

\toadd{In terms of F1 score (Fig.~\ref{fig:performance} left), DNN methods perform best on average four out of five datasets. In the remaining dataset, SMAP, ML methods obtain the best average performance.} A detailed comparison of the performance of the different methods in terms of AP \toadd{(Fig.~\ref{fig:performance} right), indicated that in terms of mean values, conventional methods outperform all families of methods in the MSL dataset and they are the second best ranked after DNN methods for the remaining datasets. Most importantly, for both F1 and AP the boxplots suggest that the differences among families of methods are relatively subtle.} 

\toadd{To confirm this hypothesis, we performed a Kruskal–Wallis test on both measurements for each group of methods to determine if the medians of the three groups of methods were equal (null hypothesis). The test was performed on each dataset. For F1, the null hypothesis was rejected only for SMD ($p < 0.05$). This indicated that there are no significant differences among the  families of methods in four of the datasets, which is consistent with what is observed in Fig.~\ref{fig:performance} (left). 

For AP, the null hypothesis was rejected on SWaT and WADI ($p < 0.05$), being consistent with what is observed in Fig.~\ref{fig:performance} (right) and indicating no significant differences among the families of methods in three out of the five datasets. We performed a post hoc analysis using a Dunn's test on SWaT and WADI every category pair. The pairwise analysis showed a significant difference ($p < 0.05$) between DNN and conventional methods in WADI and provided weak evidence to reject the null hypothesis in SWaT  between DNN and conventional methods ($0.1 < p < 0.05$). In all other pairwise tests, no significant differences were encountered.}

\begin{table*}[t]
\small
\caption{\toadd{Best and worst performing methods, in terms of F1 and AP, for each family of methodsF1 and AP per each category of methods.}}
\label{table:Summary}
\footnotesize
\centering
\begin{tabular}{ l|c|cc@{\hskip 0.26cm}|cc@{\hskip 0.26cm}|cc } 
\hline
  \multicolumn{2}{c|}{}  & \multicolumn{2}{c|}{Conventional} & \multicolumn{2}{c|}{ML} & \multicolumn{2}{c}{DNN}\\
  \cline{3-8}
 \multicolumn{2}{c|}{}  & Worst & Best &  Worst & Best &  Worst & Best \\
\hline
\multirow{2}{3em}{SWaT} & AP &  \multirow{2}{*}{VAR/SSA*} & PCA & \multirow{2}{*}{DBscan} & SVM & \multirow{2}{*}{DAGMM} & \multirow{2}{*}{USAD}\\
& F1 &  & MP  &  & LOF &  & \\ 
 
\hline
\multirow{2}{3em}{WADI} & AP & \multirow{2}{*}{VAR/SSA/MP*} & PCA &  DBscan & LOF & \multirow{2}{*}{DAGMM} &  \multirow{2}{*}{USAD} \\ 
& F1 &  & MEWMA & SVM & IF &  &  \\ 
\hline
\multirow{2}{3em}{SMD} & AP &\multirow{2}{*}{MCUSUM} & \multirow{2}{*}{ICA} &  \multirow{2}{*}{DBscan} & SVM &  \multirow{2}{*}{DAGMM} &  \multirow{2}{*}{OA}\\
& F1 &  &  & & IF &  & \\ 

\hline
\multirow{2}{3em}{SMAP} & AP & MP & MCUSUM & \multirow{2}{*}{DBscan} & SVM & \multirow{2}{*}{DAGMM} & OA \\
& F1 & VAR & ICA &   & LOF &  & USAD \\

\hline
\multirow{2}{3em}{MSL} & AP & VAR & SSA &  DBscan & SVM &  LSTM-VAE &  \multirow{2}{*}{USAD}\\ 
& F1 & MP & ICA  &  SVM & IF &  DAGMM & \\ 
\hline
\multicolumn{8}{l}{\tiny $^*$: No convergence after 10 days of execution; SVM: OC-SVM; OA: Omnianomaly}
\end{tabular}
\end{table*}

\subsection{Analysis of WADI}\label{sec:analyze}
\begin{figure*}[t]
\centering
\includegraphics[scale=0.44]{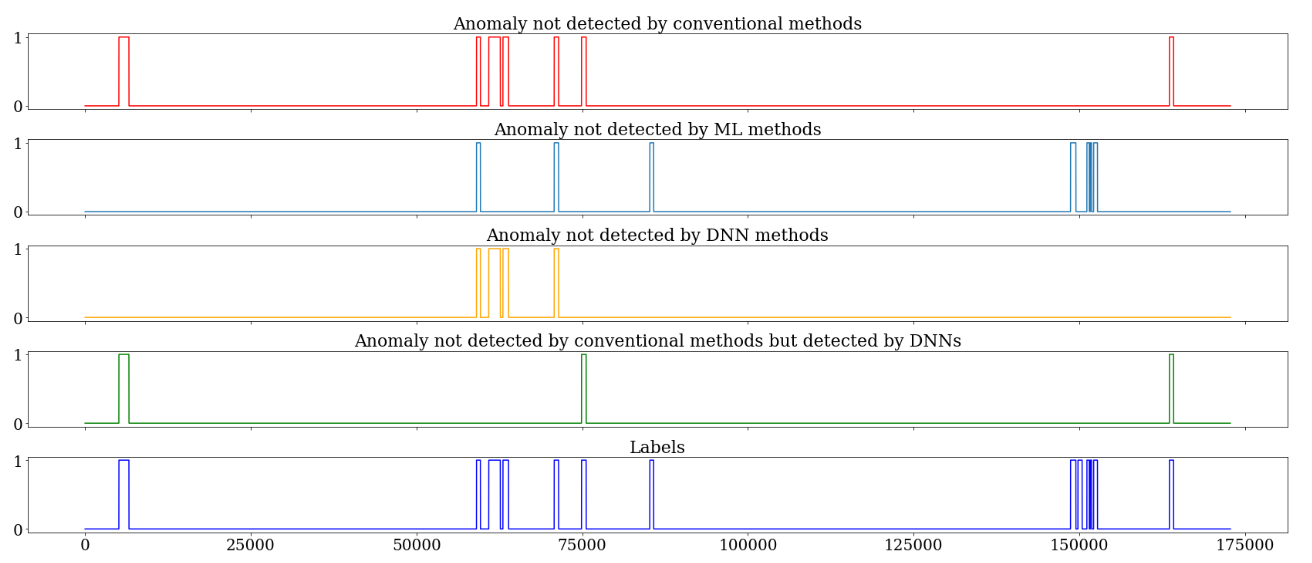}
\caption{\toadd{Analysis of WADI. }False negatives of conventional methods (top row); false negatives of ML (second); false negatives of DNNs (third); false negatives of conventional methods that are predicted by DNN (fourth) and ground truth labels (last row).}
\label{fig:wadi_ecart}
\end{figure*}

We perform a more detailed analysis on the performance of each family of methods. \toadd{We focus on the WADI dataset, as it is the only dataset where the post hoc analysis identified significant differences between the performance of DNN and the other methods.} Overall, WADI contains 14 anomalies distributed over the test set (Figure \ref{fig:wadi_ecart} bottom).

Figure \ref{fig:wadi_ecart} presents the false negatives of all the conventional methods (top), the ML methods (second row) and the DNN approaches (third row). In other words, a value of 1.0 represents an anomaly labeled as normal by a given category of methods. The fourth series (Fig. \ref{fig:wadi_ecart} fourth) presents the false negatives of the conventional approaches which are true positives for the DNNs methods. In other words, a value of 1.0 corresponds to an anomaly detected by at least one DNN algorithm but by no conventional method. 

An inspection of Figure \ref{fig:wadi_ecart} shows that four anomalies are not detected by any of the DNNs approaches, while seven anomalies are not detected by the conventional and ML models. Thus, these three anomalies explain the performance gap between the DNNs and conventional methods, which are the second best performing family in terms of AP on this dataset. 

\begin{figure*}[t]
\centering
\includegraphics[scale=0.33]{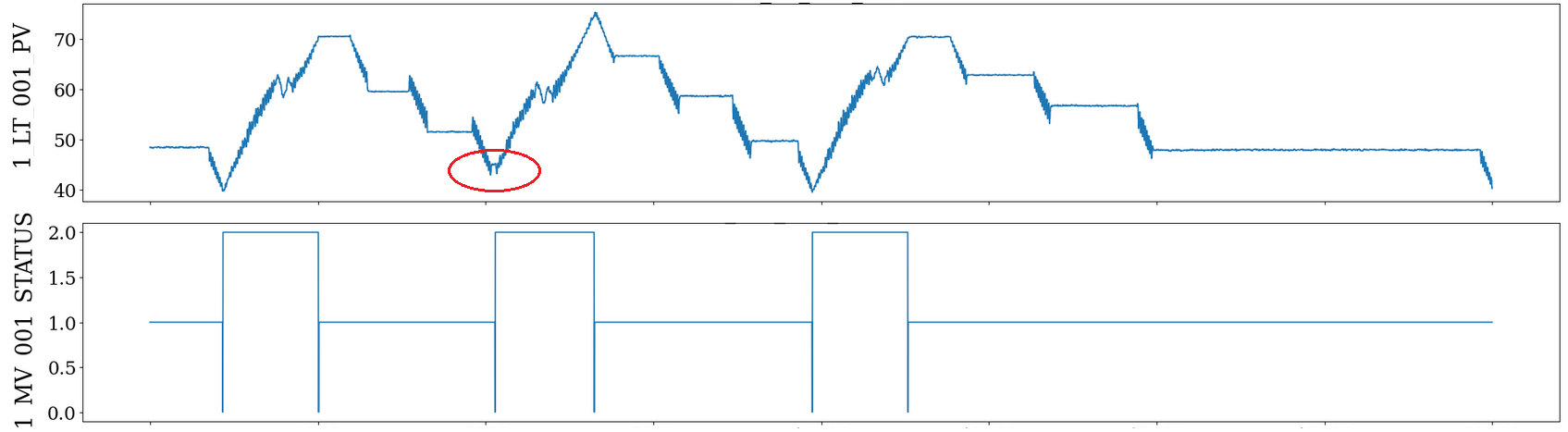}
\caption{Contextual anomaly in red due to the activation of a motorized valve (1\_MV\_001) which causes the filling of the tank (1\_LT\_001) before it reaches the switching threshold, located at 40.}
\label{fig:WADI_contextual}
\end{figure*}

Figure \ref{fig:WADI_contextual} presents a detailed view of the first anomaly from the green series. It is caused by the variate 1\_MV\_001 increasing its value to 2.0 before the variate 1\_LT\_001  has reached its minimal expected value (40). It is a contextual anomaly, as the separate behavior of each variate does not constitute an anomalous behavior on its own. The remaining two anomalies also shows that they are contextual anomalies. This suggests that the performance gap between DNNs and conventional methods comes from a better detection of contextual anomalies by DNNs approaches.

\subsection{Impact of training set size}\label{sec:size_train}
We use WADI and SWaT datasets to study the impact of the training set size on the performance of the methods. We reduce the size of the training set by keeping the most recent points, i.e. the closest to the test set. We keep 10\% , 25\% , 50\% and 75\% of the original training points. For every set, we train a model per method and assess its performance. 

\begin{figure*}[t]
\centering
\includegraphics[width=\textwidth]{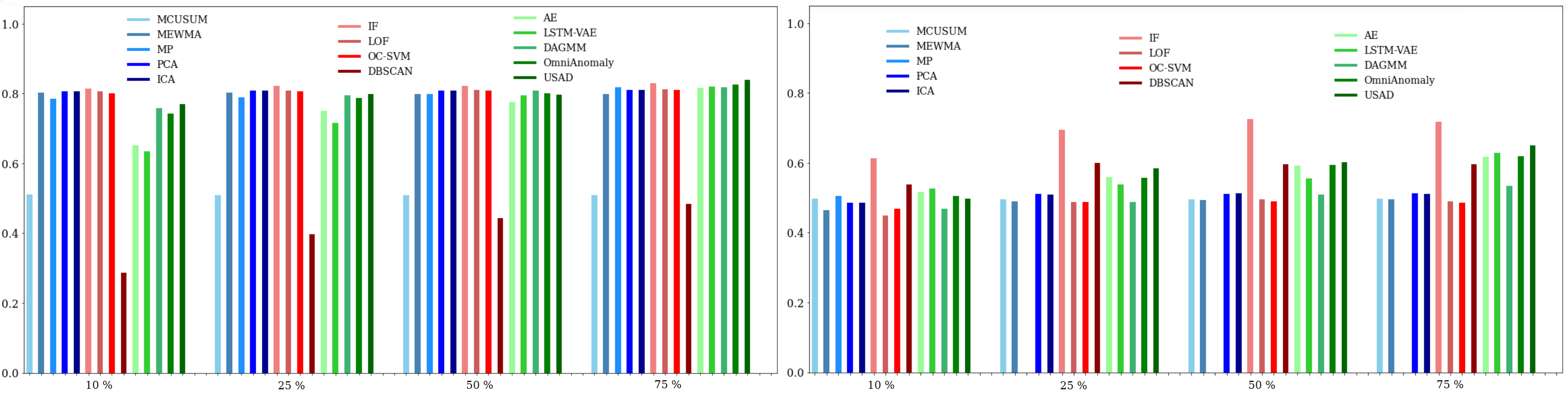}
\caption{Average precision (AP) on SWaT (left) dataset and WADI (right) datasets.}
\vspace{-0.5em}
\label{fig:roc_cut_swat}
\end{figure*}

Figure \ref{fig:roc_cut_swat} presents the AP obtained on the SWaT dataset and the WADI dataset. 
We observe that conventional methods globally obtain better results when the dataset size is less than 50\% of the original one. For example, on the SwaT dataset, MP, PCA and ICA outperform in terms of AP all ML and DNN methods when the training set is 50\% or less. Their performance is matched by OC-SVM at 50\%, and it is outperformed only when three quarters of the training dataset is retained. We also observe that the performance of conventional approaches remains relatively constant, despite changes in the training set size, meaning this has little impact in their performance. Instead, ML and DNN methods perform better when increasing the size of the training dataset, except for IF on the WADI dataset which performs slightly better when only 50\% of the dataset is kept. This can be explained by the fact that the isolation of anomalous points can be more complex when the training set is larger. In general, it is only above 50\% that DNNs approaches seem to perform better than conventional methods.

\subsection{Discussion}\label{sec:results}
\toadd{The multivariate anomaly detection performance benchmark across conventional, ML-based and DNN methods and the subsequent statistical analysis of the results (Section~\ref{sec:performance}) suggests that the performances across families of methods do not show significant differences in most of the datasets that were considered. In particular, the statistical analyses suggest that DNN methods had a clear superiority in the WADI dataset, which is consistent with our visual analysis of the obtained results (Figure~\ref{fig:performance}). 

Given the properties of WADI, a dataset containing contextual anomalies, these results suggest that} in the particular case of contextual anomalies, DNN-based methods perform better and may be necessary (Section~\ref{sec:analyze}). Indeed, contextual anomalies are by definition difficult to detect visually by experts since taken alone the series seem normal, \toadd{which may justify the need of more complex methods, such as DNNs, capable of extracting and inferring patterns from complex data. However, it is now well established that DNN techniques require large amounts of data to achieve good results. For instance, when we evaluated the performance of the different categories of methods as a function of the training set size (Section~\ref{sec:size_train}), we observed that DNN techniques observe an important drop in performance, even in those datasets where they are expected to perform best, such as WADI.}

In this work we have focused on some aspects of the performance of different types of methods, while leaving aside a detailed analysis of the computational times required by each technique. It is now relatively well established that ML and DNN-based approaches are more computationally demanding than conventional methods. This is mainly due to the fact that training time can be very expensive, especially for DNN-based methods. Conventional methods have the advantage of not requiring a training phase, which is where ML and DNN-based methods consume most of the computational resources. However, at inference time, DNN-based methods can be much faster. As an example, MP is a fast conventional approach~\cite{zhu2018}, which, however, requires to compute the distance of every new sub-sequence w.r.t. previous available time points. 

Another important observation from our study regards scalability. Some conventional methods failed to converge when the data set was too large (Table~\ref{table:Summary}). Therefore, the size of the datasets is an important criterion in the choice of a methods to use. While DNN-based methods are a more suitable choice for larger sets of data, despite their overhead in computational time during training, conventional approaches seem to be a better choice in a small data regime (Section~\ref{sec:size_train}). 

Finally, an important point to consider is also the difficulty to reproduce the results of DNN-based methods compared to the other two categories of methods. Indeed, there is a plethora of open implementations of conventional{\footnote{PYOD : https://github.com/yzhao062/pyod}} and machine learning based methods{\footnote{Scikit Learn : https://scikit-learn.org/}}, while some DNN-based approaches can be difficult to implement, sometimes no implementation is available or it is difficult to set up.

\toadd{On the basis of these observations, we consider that it is not currently possible to conclude that complex DNN methods should be the de facto choice for multivariate time-series anomaly detection. Instead, we encourage the community to reintegrate ML and conventional methods in the benchmarks to ensure that the new methods proposed improve the performance in the detection of anomalies in time series. This recommendation is inline with the observations of similar studies~\cite{makridakis2020}.}

\section{Conclusion}
In this study, we provided a comparative analysis of conventional, ML-based and DNN-based methods. We evaluated the performance of \revisiontwo{sixteen} algorithms on five publicly available benchmark datasets to understand whether the complexity provided by DNN-based approaches is necessary for anomaly detection in multivariate time series. The performance analysis did not allow us to observe a \toadd{clear and consistent} superiority of one category of methods over the others. The DNN based methods seem to perform better when the dataset contains contextual anomalies. However, this finding could only be made on one of the five datasets, so more experimentation is needed to confirm that DNN methods outperform the other categories in terms of contextual anomaly detection. We also studied the impact of the training set size on the performance of these methods. The results show that if the training set is not large enough, the conventional methods outperform the other two categories. Thus, the size of the training set is an important criterion in the choice of the category of methods to detect anomalies in multivariate time series. Some conventional methods have failed to scale on large data sets.

\toadd{ In view of all these results, we have provided the first independent evidence  for the hypothesis that much of the recent progress in time series anomaly detection may be illusory \cite{eamonn}. In \cite{eamonn}, the authors offer ``negative" evidence, noting that most benchmark datasets are unsuitable for making distinctions between algorithms. Our work offers more ``positive" evidence for the claim that deep learning has yet to prove real outperformance.} We therefore encourage the community to reincorporate the three categories of methods in the benchmarks of anomaly detection in multivariate time series. Moreover, it seems essential to us to multiply the number of datasets compared in the benchmarks in order to ensure that all eventualities are covered. For this, the community will have to obtain new real world datasets containing contextual anomalies. Indeed, the difficulty for experts to visually label contextual anomalies in multivariate time series makes it difficult to obtain test sets covering this criterion. It is then complicated to assert that DNN methods are necessary although it seems that they are able to outperform conventional approaches \toadd{when there are contextual anomalies, as long as}  the data set is large enough. 
We hope that this study will help guide researchers in their choice \toadd{and the assessment} of methods for detecting anomalies in multivariate time series. % when the type of anomalies is known in their dataset.

\toadd{
\section*{Acknowledgments}
Maria A. Zuluaga is partially funded through the 3IA Côte d'Azur Investments in the Future project managed by the National Research Agency (ANR) (ANR-19-P3IA-0002).
}

\bibliographystyle{elsarticle-num-names} 
\bibliography{cas-refs}

\end{document}